\documentclass{article}

 \usepackage[final]{corl_2018} 

\usepackage[utf8]{inputenc} 
\usepackage[T1]{fontenc}    
\usepackage{hyperref}       
\usepackage{url}            
\usepackage{booktabs}       
\usepackage{amsfonts}       
\usepackage{nicefrac}       
\usepackage{microtype}      
\usepackage{comment}

\usepackage{times}
\usepackage{xcolor}
\usepackage{subfigure}
\usepackage{soul}
\usepackage{graphicx}
\usepackage[utf8]{inputenc}
\usepackage[small]{caption}
\usepackage{wrapfig}

\title{Task-Relevant Object Discovery and Categorization for Playing First-person Shooter Games}

\author{
  Junchi ~Liang \\
  Department of Computer Science\\
  Rutgers University\\
  \texttt{junchi.liang@rutgers.edu} \\
  \And
  Abdeslam ~Boularias \\
  Department of Computer Science\\
  Rutgers University\\
  \texttt{boularias@cs.rutgers.edu} \\
}

\begin{document}

\maketitle

\begin{abstract}
We consider the problem of learning to play first-person shooter (FPS) video games using raw screen images as observations and keyboard inputs as actions. The high-dimensionality of the observations in this type of applications leads to prohibitive needs of training data for model-free methods, such as the deep Q-network (DQN), and its recurrent variant DRQN. Thus, recent works focused on learning low-dimensional representations that may reduce the need for data. This paper presents a new and efficient method for learning such representations. Salient segments of consecutive frames are detected from their {\it optical flow}, and clustered based on their feature descriptors. The clusters typically correspond to different discovered categories of objects. Segments detected in new frames are then classified based on their nearest clusters. Because only a few categories are relevant to a given task, the importance of a category is defined as the correlation between its occurrence and the agent's performance. The result is encoded as a vector indicating objects that are in the frame and their locations, and used as a side input to DRQN. Experiments on the game Doom provide a good evidence for the benefit of this approach. 
\end{abstract}
\vspace{-0.2cm}
\section{Introduction}
\vspace{-0.2cm}
The advent of deep learning has had deep impacts on several areas of artificial intelligence, and provided dramatically improved solutions to many real-world problems. The appeal of deep learning is due to the simplicity of the proposed solutions that require minimal design efforts, combined with their capability to learn complex functions. It did not take long for this thrust to reach and transform the area of reinforcement learning (RL). In the seminal work of~\citeauthor{mnih2013playing}~\citeyear{mnih2013playing}, it has been shown that a simple neural network (DQN) could be trained to play Atari video games at a human level, using raw screen images as observations and keyboard inputs as actions. 

The DQN work paved the way for several novel and improved techniques that can be categorized under the general umbrella of {\it end-to-end visual RL}. These techniques avoid the tedious process of designing features manually, and rely instead on several convolutional layers to automatically extract and learn features from sensory inputs, such as images. 

\begin{figure}[h]
  \centering
    \includegraphics[width=0.35\textwidth]{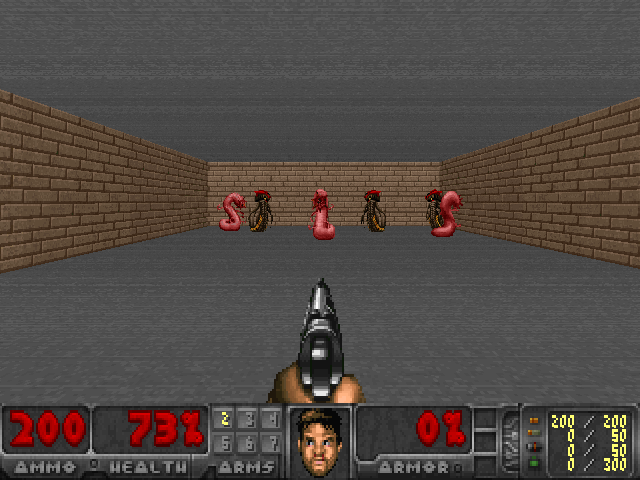}
    \includegraphics[width=0.35\textwidth]{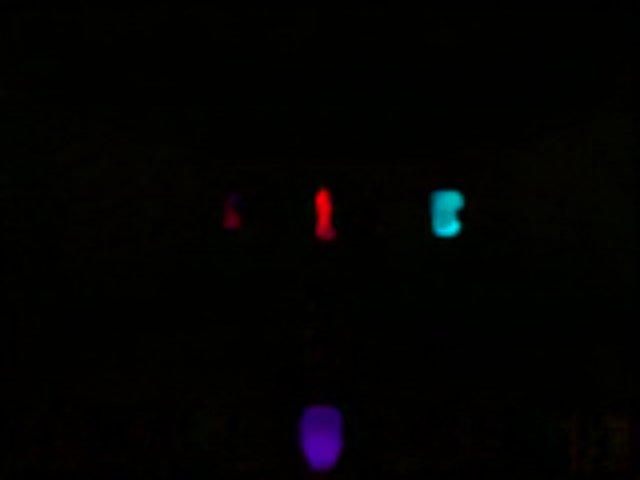}
\caption{A screenshot from the {\it Doom} game and  the corresponding detected optical flow}
    \label{fig:doom}
\end{figure}

Despite the remarkable progress made by visual RL agents in reaching human-level control and beyond, they continue to lag behind humans in terms of the number of actions they need to play before reaching a decent score. Humans are still better at quickly figuring out the effects of their actions on the objects displayed on the screen, understanding how the game works after a few trials, and utilizing a learned model for reasoning and improving their scores.

Learning predictive models from limited data and utilizing them for planning efficiently are still challenging problems. Instead, we focus in this work on augmenting the input images with channels that mark the most {\it salient} parts of the images, to speed up the learning process. {\it Saliency}, an extensively studied topic of neuroscience, refers to the property by which an object stands out relative to its neighbors. Visual saliency is an important attentional mechanism that helps learning and survival by focusing cognitive resources on the most relevant parts of a visual stimuli.

We build on recent developments in {\it optical flow} detection in videos using deep networks. Optical flows offer a robust approach for singling out and tracking moving objects, as shown in Figure~\ref{fig:doom}. Static objects, such as walls and furniture, could be detected based on their apparent motions, caused by the relative motion between an observer agent and the scene. 

The optical flow is used to segment an image into objects based on their different motions. Classical segmentation methods, such as watersheds, are used to smooth the result and remove small outliers. We then extract features of each segment from its bounding box, by using classical descriptors such as the histogram of oriented gradients (HoG). The feature vectors of all the segments detected in each frame since the agent started playing and learning are inserted into a {\it knowledge dataset}. The feature vectors in the dataset are periodically clustered.
Given a new frame, the agent uses the process described above to detect objects in the scene, and labels each object by the index of the cluster's center that is nearest to its feature vector. Therefore, the agent autonomously discovers various types of objects involved in the task and categorizes them according to their features. This acquired symbolic understanding is used alongside the raw frame to accelerate the training of a deep recurrent Q-network (DRQN).

The clusters of objects discovered by the agent could include outliers, objects that are irrelevant to the task, and noise such as fragments of the ground. Moreover, the number of categories of objects is unknown and varies depending on the task. Clearly, the inclusion of all detected objects would do more harm than good to learning. 

To focus only on types of objects that are relevant to the task, the agent performs a correlation test to find out which categories of objects its performance does depend on. 

The proposed approach is evaluated on the classical first-person shooter video game, {\it Doom}. This game provides a semi-realistic 3D world and has a better resemblance to real-world robotic tasks compared to Atari 2600. In fact, this type of FPS games could easily be seen as simulations of mobile robots navigating and acting in visually rich and highly dynamic environments. 

The empirical results reported in Section~\ref{sec:exp} show that the proposed approach clearly learns certain {\it Doom} tasks with little data compared to DRQN, while the improvement over DRQN is less pronounced on certain other tasks.

\vspace{-0.2cm}
\section{Related Work}
\vspace{-0.2cm}
\paragraph{Vision-based Reinforcement Learning:}
~\citeauthor{mnih2013playing}~\citeyear{mnih2013playing} introduced the DQN model and demonstrated its
ability to master several Atari 2600 games. A lightweight version of this model, known as A3C~\cite{pmlr-v48-mniha16} was recently introduced and shown to be computationally more efficient than DQN. Most of the recent works in this area, including ours, are built upon DQN~\cite{ArulkumaranDBB17}. For instance, \citeauthor{pmlr-v48-wangf16}~\citeyear{pmlr-v48-wangf16} presented a dueling network that represents two separate estimators, one for state value functions and one for action advantage functions.~\citeauthor{OhGLLS15}~\citeyear{OhGLLS15} proposed a different model that predicts future frames in Atari games given present frames and actions.~\citeauthor{HausknechtS15}~\citeyear{HausknechtS15} introduced the Deep Recurrent Q-Network (DRQN), where the first post-convolutional fully-connected layer of DQN is replaced with a recurrent LSTM, which is useful for dealing with partial observations that often occur in video games. Recent works started moving beyond Atari games, and focused on 3D games such as {\it Doom}~\cite{DBLP:journals/corr/KempkaWRTJ16}, which provide more realistic images and physics. Along these lines,~\citeauthor{DBLP:journals/corr/DosovitskiyK16}~\citeyear{DBLP:journals/corr/DosovitskiyK16} modified the DQN structure by adding a network that takes as inputs low-dimensional sensory data, such as health, ammunition levels, and the number of adversaries overcome. Experiments on the Doom game have shown the benefit of incorporating this side information.~\citeauthor{DBLP:conf/aaai/LampleC17}~\citeyear{DBLP:conf/aaai/LampleC17} also proposed a network for playing Doom games by training DRQN to predict {\it game features} in addition to the value function. Game features, such as the number of enemies in a given frame, are obtained from the simulator during training only. Extensions of these techniques to real-life applications, such as robotics, are also starting to take place~\cite{WatterSBR15,WahlstromSD15}.

\vspace{-0.2cm}
\paragraph{Object-based RL:}
Model-free RL techniques, such as DQN, tend to require a tremendous number of interaction data to master even simple tasks. 
Data inefficiency of model-free RL algorithms is caused by the fact that no structure of the state space or the transition model are exploited in these algorithms, which is also why they are popular and easy to use. In vision-based RL, there is a clear physical structure that could readily be  exploited. Images can be decomposed into segments of objects. Object-based representations in RL are not new, several models utilizing these representations for learning and planning had been proposed in the past~\cite{Diuk:2008:ORE:1390156.1390187}~\cite{pmlr-v32-scholz14}. 
~\cite{DBLP:journals/corr/UsunierSLC16}. More recent works focused on learning models, such as the {\it interaction networks} \cite{DBLP:conf/nips/BattagliaPLRK16}, which can reason about how objects in complex systems interact. The {\it schema network}, proposed recently by~\cite{DBLP:conf/icml/KanskySMELLDSPG17}, is an object-oriented generative physics simulator capable of disentangling multiple causes of events that occur in visual RL; it has been shown to increase transferability of skills within variations of Atari games.  These works, however, assumed that the objects are given in advance by an oracle and did not solve the problem of detecting task-relevant objects from raw sensory inputs, which is the focus of our work.   

Object-sensitive deep RL is a closely related idea proposed in~\cite{GCAI2017}. However, the approach of~\cite{GCAI2017}  uses template matching and requires that templates of objects are manually provided, while our system is fully autonomous and does not require domain-specific knowledge beyond the hyper-parameters for segmentation/clustering. Saliency detection in our method is based on statistical correlations instead of differences between Q-values as proposed in~\cite{GCAI2017}. Saliency is used to eliminate categories that are irrelevant to the agent's task and to help the agent focus on relevant objects and learn faster. In~\cite{GCAI2017}, saliency is used to visually explain the actions made. Finally, our experiments are performed on the FPS game ViZDoom, which is visually more challenging than Atari games considered in~\cite{GCAI2017}.

    \vspace{-0.25cm}
\paragraph{Self-supervised Learning of Object Categorization:}
The Gestalt principle of grouping states that humans naturally perceive objects as organized patterns and objects
~\cite{gray2006psychology}. Objects are then categorized according to various criteria, such as causal explanations~\cite{COGS:COGS451}. In computer vision, there are many works on unsupervised clustering of large sets of images based in their features as a way to categorize them. Resulting categories can be used to classify new objects by using nearest neighbors searches~\cite{DBLP:journals/corr/JohnsonDJ17}. 
We adopt a similar methodology in this work.


\vspace{-0.2cm}
\section{MDPs and Notation}
\vspace{-0.2cm}

Formally, a Markov Decision Process (MDP) is a tuple
$( S,  A, T, R, \gamma)$, where
$ S$ is a set of states and $ A$ is a set
of actions. $T$ is a transition function with $T(s,a,s')=P(s_{t+1}=s'|s_{t}=s,a_{t}=a)$ for $s,s'\in  S, a \in A$,
and $R$ is a reward function where $ R(s,a)$ is the reward
given for executing $a$ in $s$. 
A policy $\pi$ is a distribution on the action
to be executed in each state, defined as $\pi(s,a)=P(a_t=a|s_t=s)$.
The value $V^{\pi}$ of a policy $\pi$ is the expected sum of rewards
that will be received if $\pi$ is followed, i.e., $ V^{\pi}(s)=E
[\sum_{t=0}^{\infty} \gamma^t R(s_t,a_t)|s_0=s,\pi,T] $.
 The Q-value function is defined 
$Q^{\pi}(s_t,a_t) = r_t + \gamma \sum_{s_{t+1}} T(s_t,a_t,s_{t+1}) V^{\pi}(s_{t+1})$, with $r_t\stackrel{def}{=} R(s_t,a_t)$. 

In video games, the state of an agent's environment cannot be observed directly. Instead, the agent receives partial observations $o_t$, in the form of images for example. 

\begin{figure*}[t]
    \centering 
    \includegraphics[scale=0.19]{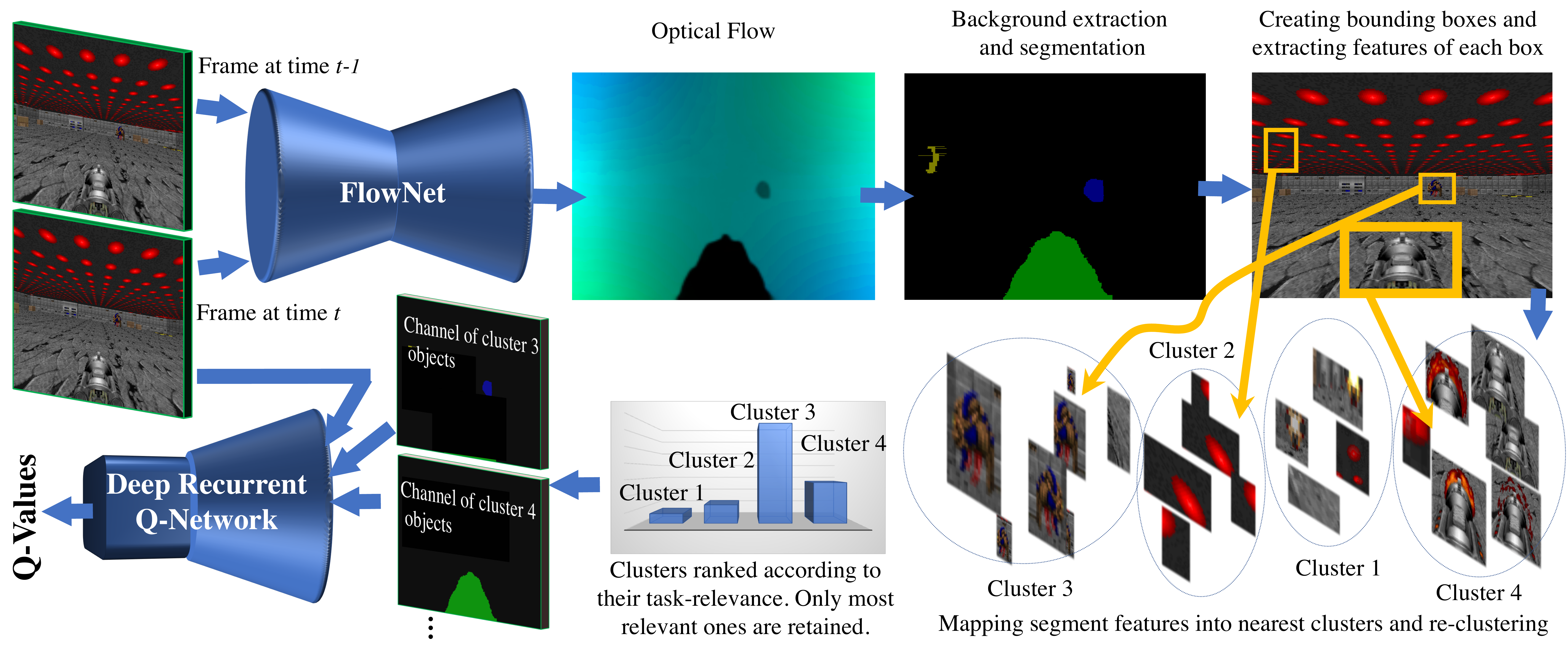} 
    \caption{Overview of the proposed approach}
 \label{fig:system}
\end{figure*}
\vspace{-0.2cm}
\section{Proposed Approach}
\vspace{-0.2cm}
\subsection{System Overview}

Figure~\ref{fig:system} shows the work-flow of the proposed system. The input is a couple of consecutive frames $o_{t-1}$ and $o_{t}$ from the video game, and the output is a vector of Q-values for the different actions. Intermediate steps consist in first computing the optical flow between the two frames (Section~\ref{Sec:optical}), detecting the background and the various objects present in the scene based on their motions relative to the camera (Section~\ref{Sec:segmentations}), mapping each segment into its nearest cluster in an ever growing knowledge dataset (Section~\ref{Sec:clustering}), inserting each segment into a channel reserved for objects from the same cluster (Section~\ref{Sec:channels}), and finally feeding a DRQN with the present frame 
$o_{t}$ and the channels of the categories to predict the Q-values (Section~\ref{Sec:prediction}). 

With an increasing probability, the action with the highest predicted Q-value is executed. The DRQN's weights are updated after each time-step, based on the observations and the received rewards. Periodically, the dataset of objects is re-clustered, leading to the potential discovery of new categories, and the relevance of each category is re-computed (Section~\ref{Sec:selection}). 

\subsection{Optical Flow Computation}
\label{Sec:optical}
An optical flow corresponds to the motion, at every pixel position, between two image frames which are taken at times $t-1$ and $t$. Formally, the optical flow can be defined as a function $f$ that maps each 
pixel $(x,y)$ at time $t-1$ to its displacement $\Delta (x,y)$ at $t$, such that the {\it intensity} $I_{t-1}$ of the image $o_{t-1}$ at $(x,y)$ is preserved, i.e. $I_{t}[(x,y)] \approx I_{t-1} [(x,y) + f(x,y)]$. There are several analytical unsupervised techniques for estimating optical flows~\cite{Lucas:1981}, by searching for regularized functions that preserve the smoothness of the objects in the images. 

Recent techniques focus on using Convolutional Neural Network (CNN) to predict the optical flow given consecutive frames as inputs. Optical flow networks are trained with labeled frames, and offer far superior quality compared to traditional unsupervised techniques. After experimenting with several architectures, we found that 
FlowNet 2.0~\cite{IlgMSKDB17} yields the best results for our purpose. Therefore, we adopt this system in the present work. 
FlowNet is trained off-line on a synthetic Flying Chairs
dataset~\cite{Aubry14}. For the fairness of the comparisons, we did not tune or re-train the FlowNet using frames from the Doom game. We expect the accuracy of the predicted flow to improve significantly if we do so, and it is easy to automatically generate labeled frames using the game engine. 

\vspace{-0.2cm}
\subsection{Optical Flow Segmentation}
\label{Sec:segmentations}
The optical flow can be represented as an HSV image with the hues as the direction of motion at each pixel and the values as the corresponding velocities. The image then can be segmented into continuous homogeneous regions that correspond to different objects, or parts of objects. 

Recent segmentation techniques utilize CNNs~\cite{ChengTW017}, but the absence of any labeled data in our task makes the results of these methods less than satisfactory. We therefore devise the following steps for segmenting the flow. 

\vspace{-0.2cm}
\paragraph{Smoothing, background extraction and segmentation:} 
We first compute the gradient $[\nabla_x f,\nabla_y f ]$ of the optical flow in the image by using finite differences. We then randomly sample a number of  pixels in the image, and use them as seeds of the {\it watershed} technique based on the gradients of the flow~\cite{Roerdink}. The result is an image segmented into regions that have similar accelerations and moving directions. Starting from the initial seeds, a segment's index is propagated from a pixel to its neighbor if the difference between their smoothed accelerations $[\nabla_x f,\nabla_y f ]$ is less than a predefined threshold $\epsilon_{seg}$. When the index of a segment is passed to a region that is already marked by another segment, the two are merged. 

Finally, segments that are too small are ignored and considered as noise, and segments that occupy most of the image are also ignored as considered as background. The absence of a detected background in an image is often an indication that the agent was turning or running too fast, which makes the optical flow blurry. Therefore, the agent simply ignores all the  segments obtained from a frame where a background could not be detected. A bounding box $seg_i$ is created around each found segment.

\vspace{-0.2cm}
\paragraph{Propagating and combining segments from different time-steps:}
The average optical flow $\bar{f}(seg_{i,t-1})$ in a given bounding box $seg_{i,t-1}$ at time $t-1$ roughly indicates the direction and velocity of the object inside. 
Therefore, one can exploit this information to predict where the object would be in frame $t$ as $seg_{i,t} = seg_{i,t-1} + \bar{f}(seg_{i,t-1})$. We found out that propagating segments through time and predicting their positions is necessary because the optical flow can be unreliable and noisy at times. Moreover, if the first-person agent does not move, then it is not possible to detect stationary objects in the scene from motion. 

To verify if an object had indeed moved into a predicted position, we compute the HoG features of its bounding box $seg_{i,t-1}$ in the RGB frame at $t-1$, and compare the features to the ones computed at $seg_{i,t}$. If the distance between the two HoG vectors is less than a predefined threshold $\epsilon_{track}$, then we consider the prediction as successful and we utilize the predicted segment $seg_{i,t}$ in the same way we utilize other segments in frame $t$. Otherwise, we assume that the object had disappeared, or changed its direction. Moreover, every segment has a predefined lifetime to avoid longterm effects of mis-segmentation. The countdown starts when a segment is first detected. If at a certain time, the predicted bounding box of a segment intersects with a newly detected segment that has similar HoG features, then the old segment dies and is replaced by the new one.

\vspace{-0.2cm}
\subsection{Clustering and Categorization}
\label{Sec:clustering}
The bounding box of a given segment is used to extract a rectangular image patch that contains the segment. The patch is resized as a square along its longest side. HoG features with nine bins and nine blocks are extracted from the grey scale values of the resized patch. The vector of HoG features is then inserted in a dataset.  Initially empty, the dataset contains all the segments detected since the training started, along with their HoG features. The dataset is clustered by using the {spectral clustering} technique~\cite{Shi:2000}. We define the distance between two segments as the cosine distance between their HoG features. We use this distance to construct the affinity matrix for the spectral clustering algorithm. Each cluster resulting from this process would correspond to a category of objects, such as persons, balls, or walls. 

The agent does not know the semantic label of the discovered categories, so it assigns a numerical label to each of them. For instance, one cluster in Figure~\ref{fig:system} contains only image patches that contain guns, and it is labeled as `` Cluster 4''.
As a side note, one could exploit this framework, in a future work, to establish a vocabulary and learn a common language between different agents.

As any other clustering algorithm, spectral clustering requires either a threshold or the number of clusters to be provided in advance. We set the number of clusters to a maximum value, allowing the agent to detect all categories of objects. Since the number of clusters may exceed the actual number of categories, different clusters could contain the same object seen from different perspectives. 

The agent uses the clusters to label new segments. Specifically, once a segment is detected in a frame, the agent inserts it in the knowledge dataset by assigning it to the nearest cluster and labeling the segment by the cluster's number.

\vspace{-0.2cm}
\subsection{Detecting Relevant Clusters}
\label{Sec:selection}
The different categories of objects discovered by the agent are sorted according to their correlation with the changes in the value function of the agent. Specifically, we define a binary random variable $c_i$ for each category of objects $i$ (i.e., each obtained cluster). At a given time-step $t$, $c_{i,t} = 1$ if a segment belonging to category $i$ is spotted in the frame $o_t$, and $c_{i,t} = 0$ otherwise. We also define a binary random variable $x_t$, such that $x_t = 1$ if $|\hat{V}^{\pi}_H(s_t) - \hat{J}(\pi)| \geq \eta$ and $x_t = 0$ otherwise, where $\hat{V}^{\pi}_H(s_t) = \sum_{i=t}^{t+H-1} \gamma^{i-t} r_t$ approximates empirically the expected sum of rewards that the agent receives during a short window of time starting at time $t$, and $\hat{J}(\pi)$ is the average value of $\hat{V}^{\pi}_H$ in the current episode. In other terms, $x_t$ is set to $1$ whenever the agent's performance in the next few steps deviates significantly from its typical performance.
 Finally, the agent computes the Pearson correlation coefficient (PCC) between each category variable $c_i$ and $x_t$. The categories that are most correlated with $x_t$ are kept and marked as task-relevant, and the other categories are simply ignored.

\vspace{-0.2cm}
\subsection{Channels of Objects}
\label{Sec:channels}
Given the list of task-relevant categories, returned by the previous step, a new ``semantic'' channel is created for each category of objects. The new channels have the same dimensions as the screen-shot frames, $360 \times 480$. Each new channel is a binary mask indicating the locations of all the objects in the frame belonging to the corresponding  task-relevant category. 
\subsection{Predicting Q-values}
\label{Sec:prediction}
\begin{figure}
  \centering
    \includegraphics[width=0.5\textwidth]{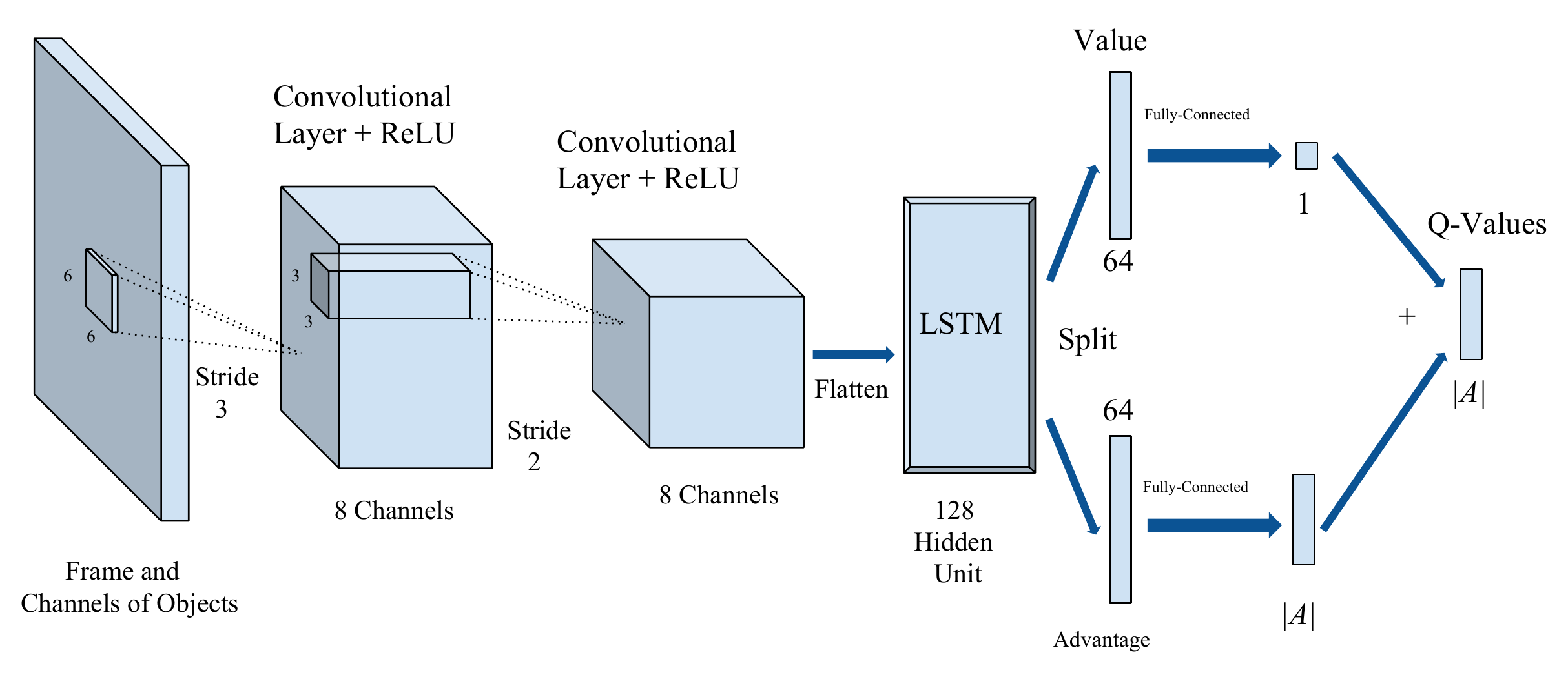}
    \vspace{-0.4cm}
\caption{Architecture of the deep recurrent Q-learning network used in the experiments}
    \label{fig:drqn}
    \vspace{-0.4cm}
\end{figure}
The final step consists in predicting the Q-value of every action, given the current frame $o_t$ and the corresponding channels of objects. 
These channels are a low-dimensional representation of the scene that help the agent pay more attention to the most salient parts of the frame and learn faster. We found out that the best performance is achieved when the  frame $o_t$ is also kept and provided as an input to the deep Q-network. The reason is that there could be important objects in the frame that are salient, but went undetected by the previous steps, or deemed irrelevant to the task. 

We optimize and adapt the DRQN design of \citeauthor{DBLP:conf/aaai/LampleC17}~\citeyear{DBLP:conf/aaai/LampleC17} to our tasks, as illustrated in Figure~\ref{fig:drqn}. The input layer is the frame and channels of the objects, and is followed by a $6 \times 6$ convolutional layer with stride $3$ and $8$ channels. This is followed by another $3 \times 3$ convolutional layer with stride $2$ and $8$ channels. Both of these two convolutional layers are followed by a ReLU \cite{NIPS2012_4824} layer. The last convolutional layer is flattened and forwarded to an LSTM with 128 hidden units. As proposed by \citeauthor{pmlr-v48-wangf16}~\citeyear{pmlr-v48-wangf16}, we split the output of the LSTM into two branches of the same size: a value branch and an action advantage branch, which are both fully-connected layers. Finally, the two outputs are combined to compute the Q-values.

\vspace{-0.2cm}
\section{Empirical Evaluation}
\label{sec:exp}
\vspace{-0.2cm}

The proposed approach is evaluated on the following four tasks, taken from the ViZDoom environment~\cite{DBLP:journals/corr/KempkaWRTJ16}: {\it Predict Position} (Task 1), {\it Defend The Line} (Task 2), {\it Defend The Center} (Task 3), and {\it The Deadly Corridor} (Task 4). 
We compare the DRQN baseline (using only game frames as inputs) to the proposed method, as well as to a variant of the proposed method that utilizes an oracle for obtaining the segments instead of relying on the optical flow and the automated segmentation pipeline. 
In the oracle variant, the segments of the objects are obtained directly from the program of the game. Comparisons to the oracle variant allow us to isolate issues that are due to the segmentation from those that are inherent to the proposed approach.

The state-of-the-art methods in ViZDoom are~\cite{DBLP:conf/aaai/LampleC17} and~\cite{DBLP:journals/corr/DosovitskiyK16}. These are both based on the D(R)QN algorithm, respectively. Hence, we chose DRQN as a baseline for comparison. We did not compare to these two methods directly because they require access to the internal state of the game engine during the training phase (such as ammunition levels, health levels, and number of adversaries present in a given frame). This is a strong requirement that goes against our driving principle of learning directly from raw sensory inputs and rewards, without any additional help. 

\begin{figure*}[h]
  \centering
   \subfigure[{\scriptsize Task 1: Predict Position}]
 {
    \includegraphics[width=0.24\textwidth]{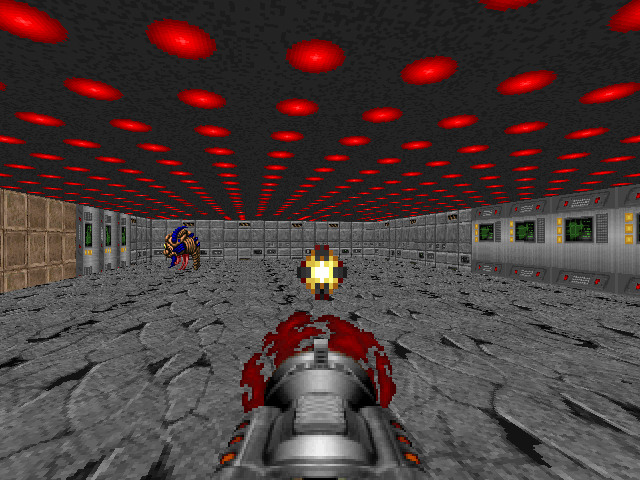}
    }
    \hspace{-0.3cm}
     \subfigure[{\scriptsize Task 2: Defend The Line}]
 {
    \includegraphics[width=0.24\textwidth]{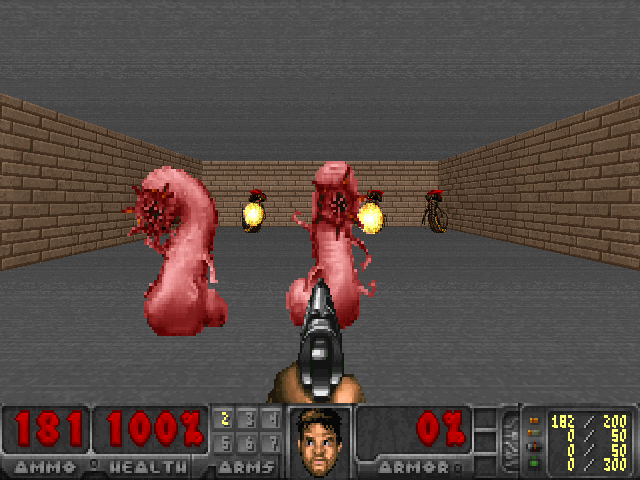}
 }
	\hspace{-0.3cm}
\subfigure[{\scriptsize Task 3:  Center Defense}]
 {
    \includegraphics[width=0.24\textwidth]{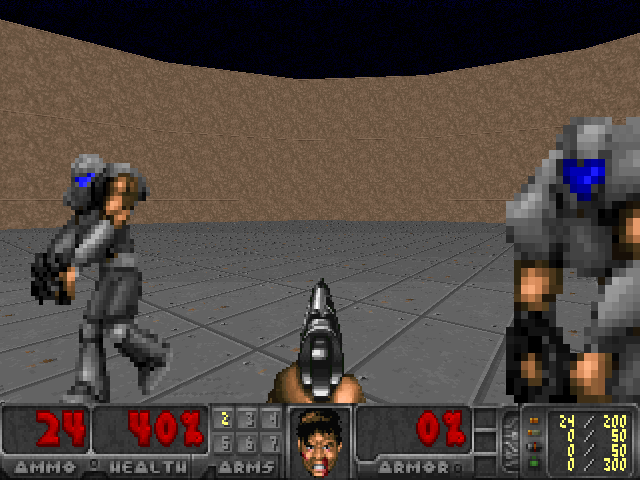}
 }
     \hspace{-0.3cm}
     \subfigure[{\scriptsize Task 4: Deadly Corridor}]
 {
     \includegraphics[width=0.24\textwidth]{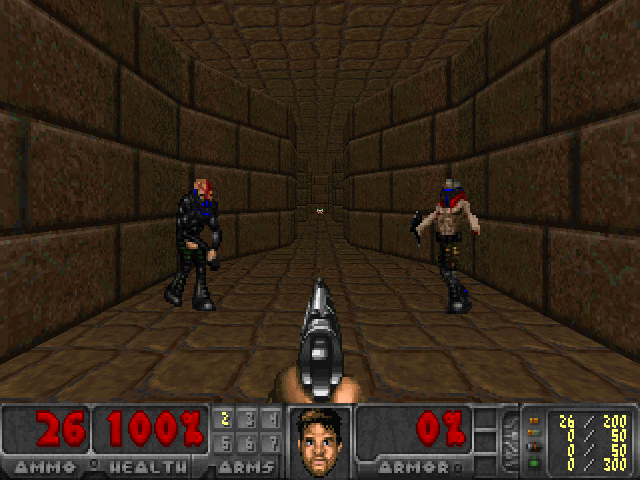}
 }
    \subfigure[\small Results of Task 1]
    {
\includegraphics[width=0.4\textwidth]{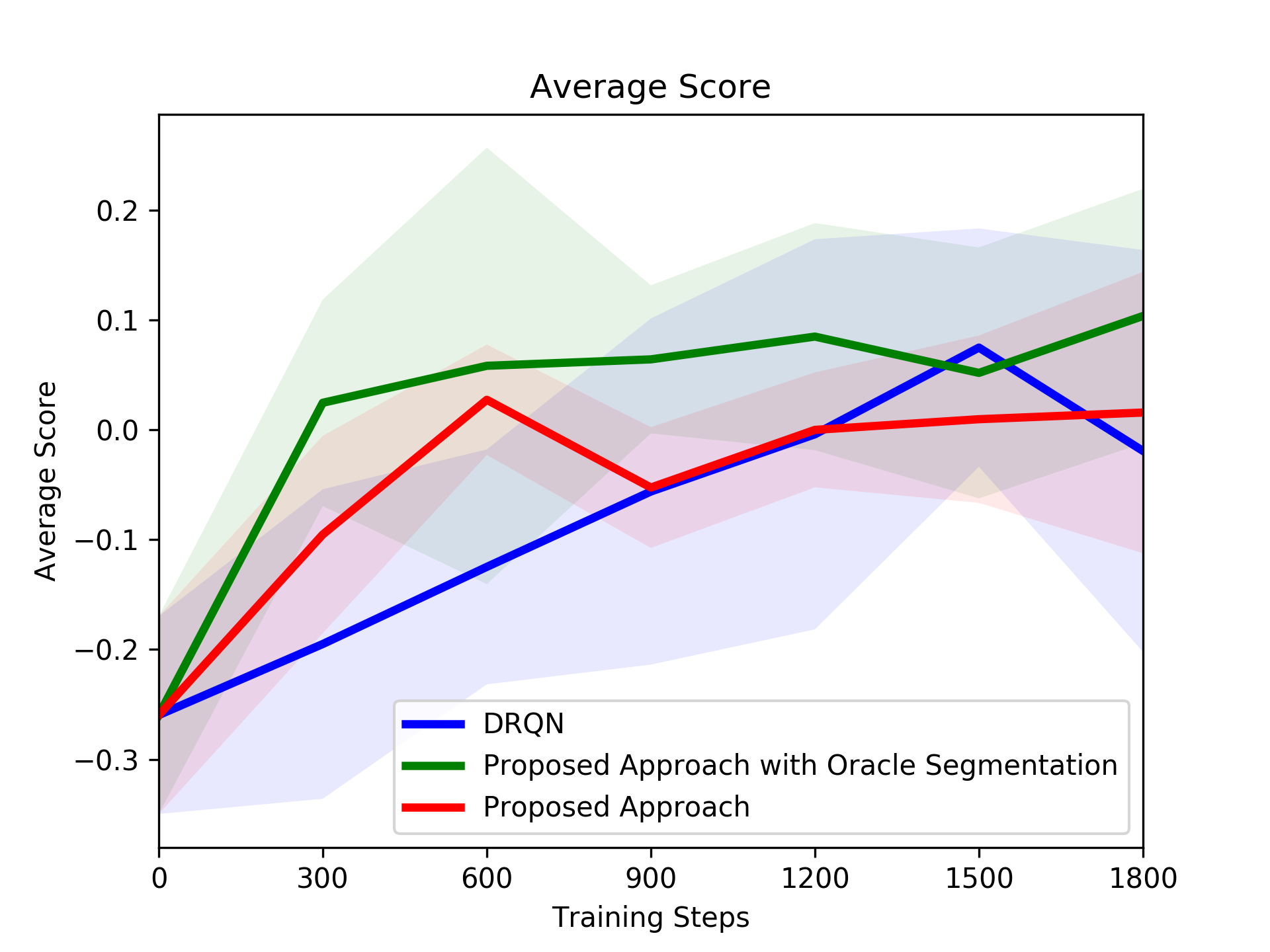}
}
     \hspace{-0.8cm}
      \subfigure[\small Results of Task 2]{
\includegraphics[width=0.4\textwidth]{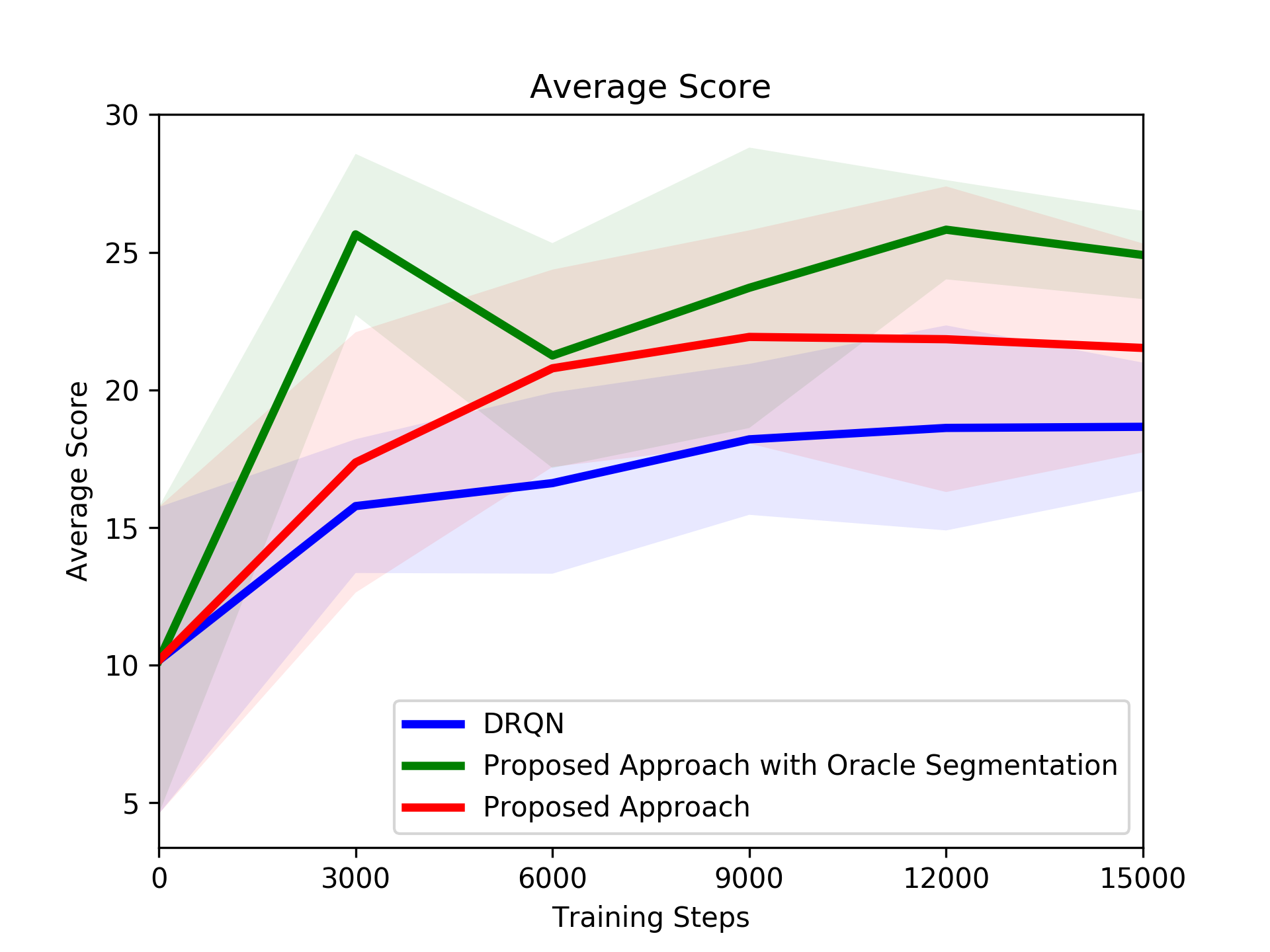}     
}
	\subfigure[\small Results of Task 3]{
\includegraphics[width=0.4\textwidth]{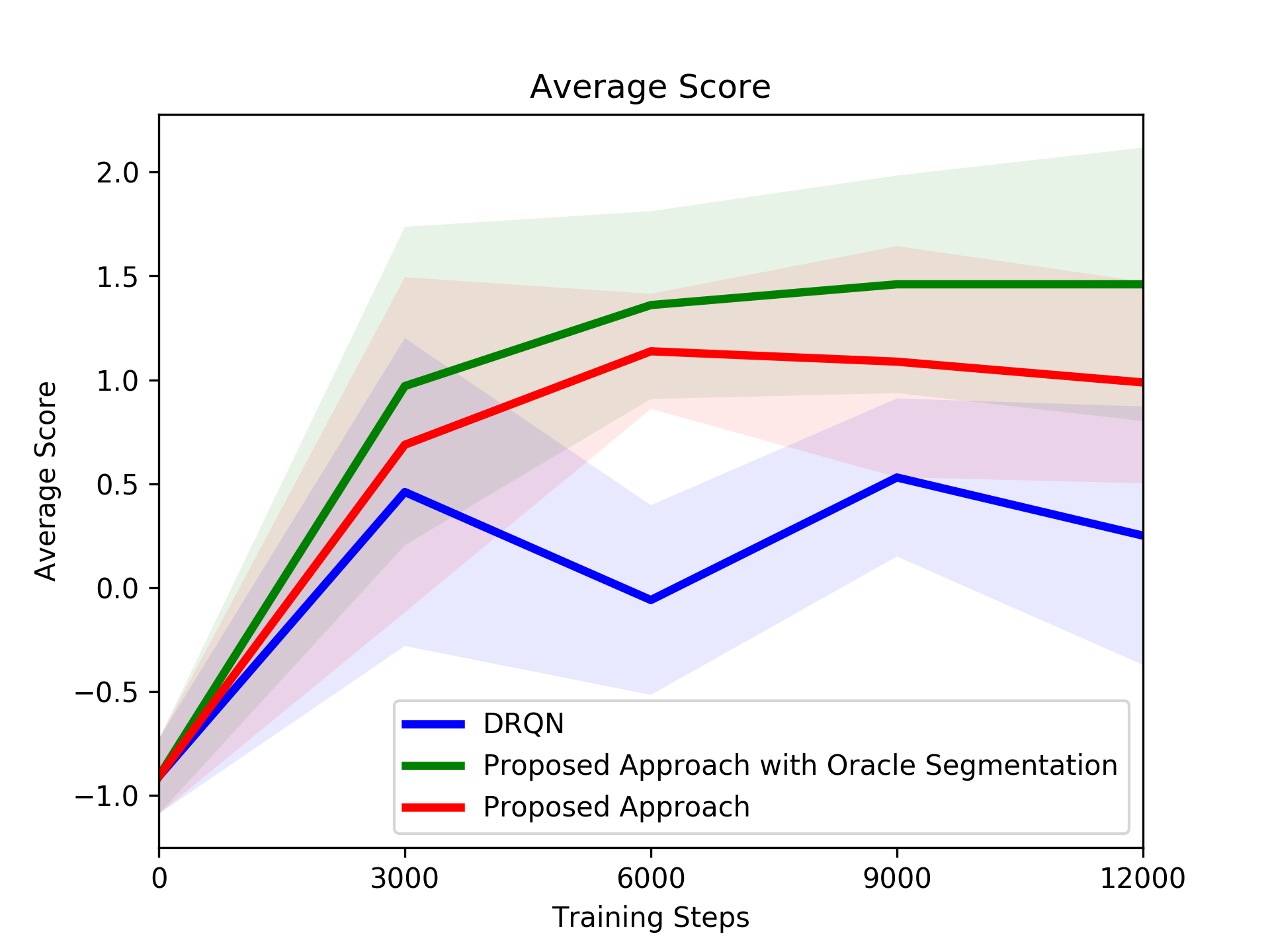} 
}
     \hspace{-0.8cm}
\subfigure[\small Results of Task 4]{
\includegraphics[width=0.4\textwidth]{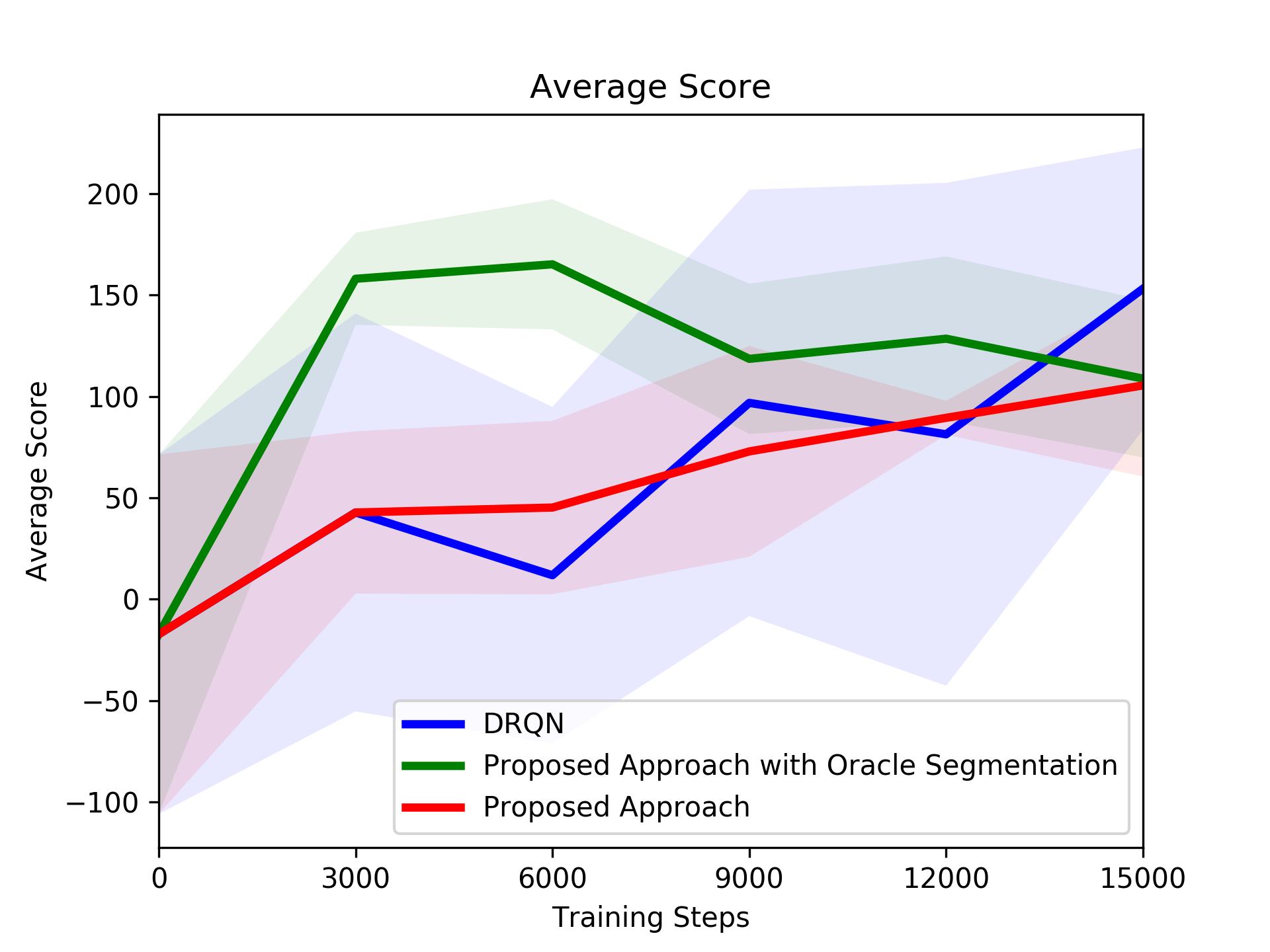}
 }
 \caption{Screenshots from the tasks considered in the experiments and the corresponding learning curves. The results are averages of five independent and random runs. The proposed approach clearly outperforms DRQN in two tasks, and performs equally in two others. An explanation of this behaviour is provided in the discussion.}
 \label{fig:results}
\vspace{-0.4cm}
\end{figure*}
\vspace{-0.25cm}
\subsection{Tasks}
\paragraph{Task 1 ({\it Predict Position}):} 
The goal is to shoot a moving object in a room. The object is randomly spawned on the opposite side of the room. The agent is equipped with a rocket launcher that can fire a missile that has a long delay between firing and hitting. When the missile hits the object, the agent gets a reward of +1. An episode times out after 300 steps.
\vspace{-0.25cm}
\paragraph{Task 2 ({\it Defend The Line}):}   The agent fights against two types of adversaries. The agent gets a reward of +1 when it shoots an adversary. It gets a reward of $-10^{-5}$ when it is hit and -1 when it dies. 
\vspace{-0.25cm}
\paragraph{Task 3 ({\it Defend The Center}):} The agent fights against adversaries that would come near and attack in a circular room. The agent is equipped with a gun and 26 ammunitions. The task's reward and episode length is similar to Task 2.
\vspace{-0.25cm}
\paragraph{Task 4 ({\it Deadly Corridor}):} The agent navigates along a passageway. The reward is the distance to the goal. There are pairs of adversaries shooting at the agent. Episodes end when the agent dies or it reaches the goal or when a timeout occurs.

\vspace{-0.25cm}
\subsection{ Parameter Settings}
\vspace{-0.05cm}
\paragraph{Segmentation:} The $480 \times 640$ screen frames are resized into $320 \times 480$ frames for the segmentation pipeline. In all tasks, we set $\epsilon_{seg}$ to $0.02$. When applying the watershed technique for the background extraction, only regions larger than $13\%$ of the image are considered as background. If less than half of a frame is labeled as a background, the segmentation of that frame is skipped because the agent is moving too fast. We set $\epsilon_{track}$ to $0.85$ for propagating segments through different time-steps. In tasks 1 and 2, the lifetime of detected segments is six frames, it is set to 12 in tasks 3 and 4.
\vspace{-0.25cm}
\paragraph{Learning:} We use the same $\epsilon$-greedy exploration strategy for the three methods. Exploration probability $\epsilon$ is initialized to $1.0$ and decreased by $15\%$ every $300$ iterations in Task 1 and every $3000$ iterations in Tasks 2 and 3, and 4. $\epsilon$ remains at $0.1$ when it reaches this probability. The discount factor $\gamma$ is set to $0.99$. The LSTM is unrolled through ten steps in all tasks, as proposed by \cite{DBLP:conf/aaai/LampleC17}. The size of a training batch is set to $32$ sequences of length ten, $\{(s_k,a_k,r_k,s_{k+1}), \dots (s_{k+9},a_{k+9},r_{k+9},s_{k+10})\}$. 

\begin{wrapfigure}{r}{0.3\textwidth}
\vspace{-0.5cm}
{\scriptsize
\begin{tabular}{p{1.58cm}p{0.3cm}p{0.3cm}p{0.3cm}}
\hline
      Category          & {Det.} & {IoU} & {Cat.}\\
\hline \hline
{\it Zombieman}& 25$\%$ & 35$\%$ &   $82\%$\\
{\it ShotgunGuy}& 15$\%$ & 33$\%$ &   $81\%$\\
{\it ChaingunGuy}& 17$\%$ & 2$\%$  &   $87\%$\\
{\it DoomPlayer1}& 23$\%$ & 6$\%$ &   $53\%$\\
{\it MarineChain}& 34$\%$ & 25$\%$  &  $1\%$\\
{\it DoomPlayer2}& 47$\%$ & 11$\%$  &  $37\%$\\
{\it DoomImp}& 29$\%$ & 12$\%$  &   $45\%$\\
{\it DoomImpBall}& 29$\%$ & 22$\%$  &   $38\%$\\
{\it Demon}& 69$\%$ & 17$\%$  &   $78\%$\\
{\it DoomPlayer3}& 27$\%$ & 10$\%$  &  $24\%$\\
{\it Cacodemon}& 75$\%$ & 23$\%$  & $84\%$\\
{\it Rocket}& 23$\%$ & 13$\%$  &  $36\%$\\
{\it DoomPlayer4}& 5$\%$ & 7$\%$  & $0\%$\\
\hline
 \end{tabular}}
\caption{\footnotesize Average detection rate (Det.), intersection over union (IoU), and categorization accuracy (Cat.) of detected segments belonging to  task-relevant categories in the four tasks.  }
\label{table}
\vspace{-0.5cm}
\end{wrapfigure}

\subsection{Discussion}

We start by evaluating the optical flow-based segmentation and the categorization parts of the proposed system. Table~\ref{table} summarizes the results of these evaluations for the categories of objects that are most relevant to the four tasks. An object is successfully detected in a frame if a segment belonging to it is returned. Despite the low detection rate, the agent propagates over time detected segments and predicts their positions in next frames based on their observed motion using the optical flow. Table~\ref{table} also shows that relatively medium or small portions of objects are detected on average, but this seems enough to extract HoG features that help clustering and categorizing the segments. A categorization is considered accurate when a segment is assigned to a cluster containing mostly similar segments (based on its HoG distance). The agent systematically ignores categorization outcomes whenever the nearest cluster to a segment is considered too far, which makes it more robust to outliers.

It is important to note that the agent does not need to have a high detection rate to know where objects are and to play well. Thanks to the inferred directions and velocities using the optical flow, the agent can keep track of the positions of objects for several seconds even if they are successfully detected only once every ten or so frames.

Figure~\ref{fig:results} shows the average scores per time-step obtained by the three methods, as functions of the number of time-steps in training. These results are obtained by performing $20$ episodes of testing after every $300$ training time-steps for Task 1, and after every $3000$ time-steps for Tasks 2, 3, and 4. The results are averaged over five independent experiments. 

As expected, the oracle variant of our approach outperforms the fully autonomous approach in all tasks because it gets a perfect segmentation directly from the game's API. The  performance of the  fully autonomous method is mixed, but seems to be always better than or equal to the DQN baseline. In fact, the results can be split into two categories, as follows.

\noindent {\bf a) The proposed method outperforms DRQN}. This is the case of Tasks 2 and 3. In these two tasks, the proposed method not only learns faster at the beginning, but also seems to converge to a higher value than DRQN. This outperformance is explained by the strong correlation between the detected object categories and the reward that the agent receives. Indeed, the appearance of adversaries in the scene often leads to an immediate decrease in the expected cumulative reward in the next time-steps if the agent does not react well. Moreover, shooting at the adversaries has an immediate positive reward, which creates a strong correlation between these objects and the value function. The improvement brought by our method is slightly less pronounced in Task 2 compared to Task 3, because there are more categories of objects in Task 3, and these categories affect the agent differently.

\noindent {\bf b) The proposed method performs similarly to  DRQN}.
This is the case of Tasks 1 and 4, where despite an accelerated improvement in the early time-steps, DRQ catches up and seems to perform equally well. These tasks are particularly challenging, because of the long delay between firing a missile and the moment it hits the target and the agent receives a reward in Task 1, and the fact that the reward in Task 4 is related to the distance between the agent and its goal. Therefore, it is not easy to correlate the visual stimuli with the performance of the agent. Moreover, the target being too far ahead, it is often hard to capture it from its optical flow. This is perhaps the reason why the oracle approach seems to outperform DRQN in these tasks as well.

\vspace{-0.25cm}
\vspace{-0.2cm}
\section{Conclusion}
\vspace{-0.25cm}
We presented an approach for improving the data-efficiency of visual RL methods by  detecting salient parts in the images and categorizing them based on how their occurrences correlate with the changes in the agent's received cumulative rewards. Experiments on FPS tasks show the advantage of the proposed approach in cases where there is a strong correlation between the perceived visual stimuli and the received rewards. This method is an example of works where combining novel deep learning architectures with classical non-parametric methods could result in an improved performance.
In a future work, we will explore temporal causality models that take into account longer delays in the perceived events. It would be interesting to evaluate the proposed method on a mobile robot, where salient objects can be detected from their relative motions with respect to the robot. 

\bibliography{ijcai18}

\end{document}